\documentclass[sigconf]{acmart}

\usepackage{booktabs} 
\usepackage{multirow}
\usepackage{subfigure}
\usepackage{verbatim}
\usepackage{CJKutf8}
\usepackage{flushend}

\usepackage{makecell}
\usepackage[utf8]{inputenc}

\copyrightyear{2022}
\acmYear{2022}
\setcopyright{rightsretained} 
\acmPrice{15.00}

\begin{document}

\title{Unified and Effective Ensemble Knowledge Distillation}


\author{Chuhan Wu$^1$, Fangzhao Wu$^2$, Tao Qi$^1$, Yongfeng Huang$^1$}

\affiliation{%
  \institution{$^1$Department of Electronic Engineering \& BNRist, Tsinghua University, Beijing 100084 \\ $^2$Microsoft Research Asia, Beijing 100080, China}
} 
\email{{wuchuhan15,wufangzhao,taoqi.qt}@gmail.com,yfhuang@tsinghua.edu.cn}

\begin{abstract}

Ensemble knowledge distillation can extract knowledge from multiple teacher models and encode it into a single student model.
Many existing methods learn and distill the student model on labeled data only.
However, the teacher models are usually learned on the same labeled data, and their predictions have high correlations with groudtruth labels.
Thus, they cannot provide sufficient knowledge complementary to task labels for teaching student.
Distilling on unseen unlabeled data has the potential to enhance the knowledge transfer from the teachers to the student.
In this paper, we propose a unified and effective ensemble knowledge distillation method that distills a single student model from an ensemble of teacher models on both labeled and unlabeled data.
Since different teachers may have diverse prediction correctness on the same sample, on labeled data we weight the predictions of different teachers according to their correctness.
In addition, we weight the distillation loss based on the overall prediction correctness of the teacher ensemble to distill high-quality knowledge.
On unlabeled data, the disagreement among teachers is an indication of sample hardness, and thereby we weight the distillation loss based on teachers' disagreement to emphasize knowledge distillation on important samples.
Extensive experiments on four datasets show the effectiveness of our proposed ensemble distillation method.

\end{abstract}


\maketitle

\section{Introduction}

\textit{Two heads are better than one}. Instead of using a single model, leveraging an ensemble of multiple models is a simple yet effective strategy that can usually boost the accuracy~\cite{dietterich2000ensemble}.
Ensemble techniques have empowered various classification~\cite{tsoumakas2007random} and regression~\cite{mendes2012ensemble} tasks.
However, different from traditional shallow and small ensemble models such as boosting~\cite{schapire1999brief} and random forest~\cite{ho1995random}, it is difficult to use ensembles of big models (e.g., BERT~\cite{devlin2019bert}) for inference in low-latency systems due to the huge computational cost~\cite{xu2020improving}.

Knowledge distillation from multiple teachers aims to obtain a strong student that inherits most performance of the teacher ensemble without increasing the inference computational cost~\cite{zhu2018knowledge,cho2019efficacy}.
This paradigm is known as ensemble knowledge distillation~\cite{allen2020towards}.
There are many prior studies on ensemble knowledge distillation~\cite{liu2019improving,wang2020distilling,walawalkar2020online,wu2021one,kang2020ensemble}.
Most methods distill the student model on labeled data, and a core problem they addressed is assigning different ensemble weights for different teachers.
For example, \citet{chebotar2016distilling} proposed to first search optimal constant weights for combining teacher models' outputs that yield the best accuracy, and then distill a student from the ensemble soft labels.
\citet{du2020agree} proposed an adaptive weighting method by using different teacher ensemble weights for different samples that minimize the classification loss.
These methods usually equally regard the importance of different labeled samples in knowledge distillation, which may be suboptimal because teachers' predictions on different samples may have different helpfulness.
In addition, they usually learn and distill the student model on the same labeled samples from which the teacher models are trained.
However, the teacher models' predictions on these samples do not necessarily reflect their real prediction patterns on the overall data distribution due to their memory of labels~\cite{krishnan2020improving}.
Thus, it is insufficient to transfer knowledge on labeled data only.
\begin{figure*}[!t]
  \centering 
      \includegraphics[width=0.85\linewidth]{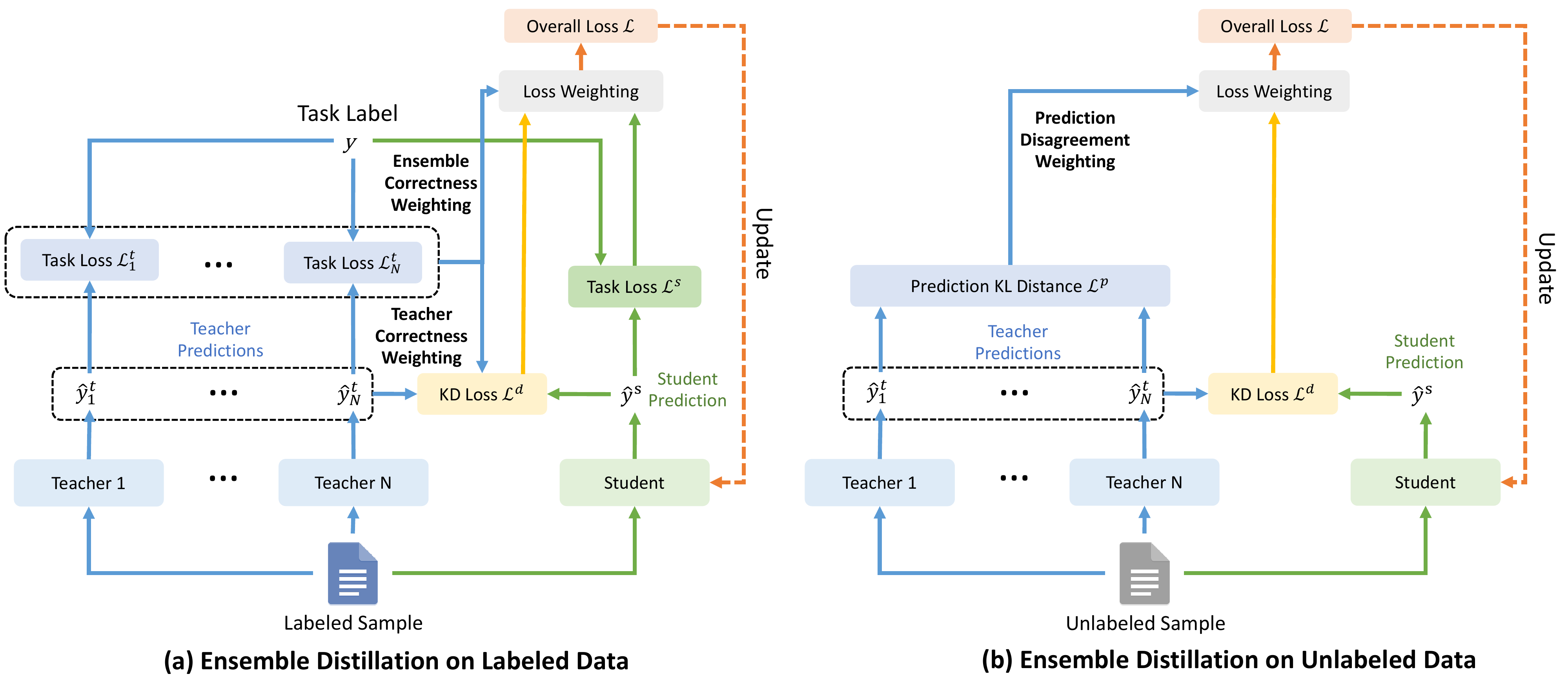} 
  \caption{The unified ensemble distillation framework of UniKD.}\label{fig.model} 
\end{figure*}

There are a few approaches for ensemble distillation on unlabeled data~\cite{sun2019token,sui2020feded,lin2020ensemble,gong2021ensemble}.
For example, \citet{radosavovic2018data} proposed to apply teacher models to unlabeled data with different augmentation methods to obtain ensemble predictions for student teaching.
\cite{li2019fedmd,sui2020feded} used the average predictions of multiple models on an unlabeled dataset as the teaching signals for distilling individual models.
However, these methods cannot distinguish between important and uninformative unlabeled samples, which is critical for fully distilling teachers' knowledge. 
In fact, the teacher models may have different prediction disagreement on different unlabeled samples, and it is important to actively learn more on borderline samples with strong disagreement to improve the prediction quality of the student model~\cite{dekel2012selective}.

In this paper, we propose a unified ensemble knowledge distillation method named UniKD, which can distill a high-quality student model from multiple teacher models on both labeled data and unlabeled data in a unified way.
Since different teachers have different prediction correctness, on labeled data we weight teachers' soft labels according to their losses on each sample to encourage the student to learn more from the accurate teachers.
To further help distill high-quality knowledge, we weight the knowledge distillation loss on each labeled sample based on the average loss of teachers, which can enforce the student to learn more from the task label rather than teachers if teachers' error is high.
On unlabeled data, since there is no task label to measure prediction correctness, we average the soft label predictions of teachers for knowledge distillation.
To help distill knowledge more effectively on important samples, we use the disagreement of teachers' predictions on each sample to weight the unlabeled distillation loss by emphasizing the samples on which a single teacher model has high variance and low confidence.
Extensive experiments on four benchmark datasets show that UniKD enables the student model to beat any single teacher model with a large margin, and can outperform many baseline methods.

\section{Methodolody}\label{sec:Model}

We then introduce the details of UniKD.
Its knowledge distillation frameworks on labeled and unlabeled data are shown in Fig.~\ref{fig.model}.\footnote{We consider knowledge distillation from the output labels rather than features~\cite{park2019feed}, because it does not add any requirement to the model architectures and hyperparameters.}
We discuss each of them in the following sections.

\subsection{Ensemble Distillation on Labeled Data}

Distilling the student model on labeled data can prevent it from overfitting task labels~\cite{yuan2020revisiting}.
Thus, we also consider labeled data-based ensemble distillation, as shown in Fig.~\ref{fig.model}(a).
Its has a two-level weighting mechanism based on the prediction correctness of each individual teacher and all teachers.
Assume there is an ensemble of $N$ teacher models to teach the student.
We denote the soft labels predicted by the student and $i$-th teacher on a training sample as $\hat{y}^s$ and $\hat{y}^t_i$, respectively.
We use the groundtruth label $y$ of this sample to compute the task losses of teachers and student on this sample, which is denoted as $\mathcal{L}^t_i$ for the $i$-th teacher and $\mathcal{L}^s$ for the student.
Since the teacher model with high prediction errors can be misleading, we first weight different teachers' soft labels of each sample based on their prediction losses.
More specifically, the ensemble prediction $\hat{y}^t$ of teacher models on a sample is computed as follows:
\begin{equation}
    \hat{y}^t = \sum_{i=1}^N \frac{\mathcal{L}^t_i }{\sum_{j=1}^N \mathcal{L}^t_j}\hat{y}^t_i.
\end{equation}
This formulation means that a higher loss on a specific sample yields lower importance in the prediction ensemble.

Next, we use the ensemble prediction $\hat{y}^t$ to teach the student.
We use a crossentropy loss $\mathcal{L}^d$ to regularize the student model to make similar predictions with the ensemble predictions, which is formulated as follows:
\begin{equation}
\mathcal{L}^d= -\sum_{i=1}^C\hat{y}^t[i]\log(\hat{y}^s[i]),
\end{equation}
where $[i]$ means the $i$-th element of the soft label, and $C$ is the number of classes.
The student model also learns from the label $y$ in the target task using the task loss $\mathcal{L}^s$.
Since on different samples the teachers' predictions have different qualities, it is important to dynamically adjust the relative importance of the supervision from the teacher ensemble and the task labels.
Thus, we weight the distillation loss and task loss when combining them into an overall loss $\mathcal{L}$ as follows:
\begin{equation}
    \mathcal{L}=\frac{1}{1+\sum_{i=1}^N\mathcal{L}^t_i/N}\mathcal{L}^d + \mathcal{L}^s.
\end{equation}
In this way, the student model learns more from the task label when the teacher models' predictions are inaccurate, which can facilitate high-quality knowledge transfer.

\subsection{Ensemble Distillation on Unlabeled Data}

Since knowledge distillation on optimized labeled data may not fully distill teachers' knowledge, we also consider  ensemble knowledge distillation on unseen unlabeled data to enhance knowledge transfer, as shown in Fig.~\ref{fig.model}(b).
Since there are no task labels to evaluate the teachers' predictions, we directly average their predicted soft labels as the ensemble prediction $\hat{y}$, which is further used to compute the knowledge distillation loss $\mathcal{L}^d$ in the same way.
However, the teachers' predictions on different unlabeled samples may have different disagreements~\cite{kao2021specific}.
If different teachers have very consistent predictions on a sample, it means that every single teacher model  classifies this sample correctly/incorrectly.
In this case, the student model should learn less from the teacher because ensemble cannot improve the performance on this sample.
On the contrary, a strong disagreement on a sample means that a single model's prediction has high variance on this sample, and model ensemble can help reduce the uncertainty.
Thus, the knowledge distillation intensity on this sample should be strong to better encourage the student to mimic the teacher ensemble.
Motivated by the above observations, we propose to weight the knowledge distillation loss on unlabeled data based on the disagreement among teachers.
To measure the disagreement of teachers' predictions, we use the average Kullback–Leibler (KL) divergence between all pairs of teacher predictions.
The disagreement score $\mathcal{L}^p$ on a sample is calculated as follows:
\begin{equation}
    \mathcal{L}^p=\frac{1}{N(N-1)}\sum_{i\neq j}{KL(\hat{y}^t_i, \hat{y}^t_j)},
\end{equation}
where a higher score indicates that teachers' predictions are more diverse.
We use this score to further weight the knowledge distillation loss, and the overall loss $\mathcal{L}$ is formulated as follows:
\begin{equation}
    \mathcal{L}=(1+\lambda \mathcal{L}^p)\mathcal{L}^d,
\end{equation}
where $\lambda$ is a hyperparameter that controls the influence of teacher disagreement on the loss function.
When both labeled and unlabeled data are available, we combine the knowledge distillation losses on all samples.
By optimizing the distillation loss, the student model can be tuned by supervision signals, meanwhile fully inheriting the knowledge encoded by the multiple teachers.

\section{Experiments}\label{sec:Experiments}

\subsection{Datasets and Experimental Settings}

We conducted experiments on four benchmark datasets, including MNLI~\cite{williams2018broad}, QNLI~\cite{rajpurkar2016squad}, QQP\footnote{https://data.quora.com/First-Quora-Dataset-Release-Question-Pairs} and SST-2~\cite{socher2013recursive}, which are taken from the GLUE~\cite{wang2018glue} benchmark.\footnote{They are selected because they have relatively large sizes and are easy to split part of data as unlabeled dataset.}
The statistics of these datasets are summarized in  Table~\ref{table.dataset}.
On all datasets, we use half of the training data as labeled data, and regard the rest as unlabeled data by removing their labels.
Following~\cite{bao2020unilmv2}, we report the results on the dev set because test labels are not released.

\begin{table}[h]
\centering

	\caption{Statistics of the datasets used in our experiments.}\label{table.dataset}

\begin{tabular}{lcccc}
\Xhline{1.5pt}  
        & \textbf{MNLI} & \textbf{QNLI} & \textbf{QQP} & \textbf{SST} \\ \hline
\#train & 393k          & 105k          & 364k         & 67k          \\
\#val   & 20k           & 5.5k          & 40k          & 872          \\
\#tet   & 20k           & 5.5k          & 391k         & 1.8k         \\ \hline \Xhline{1.5pt}            
\end{tabular} 

\end{table}

In our experiments, we use Adam~\cite{kingma2014adam} as the model optimizer. 
The learning rate is 1e-5 and the batch size is 16.
Without loss of generality, the student and teacher models have the same architecture and sizes.\footnote{Our approach can be directly applied to the scenarios where the teachers and the student have different architectures and sizes.}
The value of $\lambda$ is 10 or 15 (see experiments).
Following~\cite{liu2019roberta}, we repeat each experiment five times, and the five independent models are used for ensemble.
We sample 10\% of labeled training data as validation data for tuning hyperparameters, and then use all labeled data for model training and test.
We use classification accuracy as the metric, and we report the average results of student model in another five repeated experiments.

\begin{table}[t]
\caption{Performance of different distillation methods.}\label{table.kd}
\begin{tabular}{lcccc}
\Xhline{1.5pt}  
\multicolumn{1}{c}{\textbf{Methods}} & \textbf{MNLI} & \textbf{QNLI} & \textbf{QQP} & \textbf{SST} \\ \hline
BERT (single)                        & 82.9          & 88.8          & 88.1         & 91.6         \\
BERT (ensemble)                      & 84.3          & 90.6          & 89.8         & 93.7         \\
BERT+KD-Labeled                              & 83.3          & 89.3          & 88.4         & 91.9         \\
BERT+AE-KD                           & 83.5          & 89.5          & 88.6         & 92.2         \\
BERT+KD-Unlabeled                             & 83.7          & 89.9          & 89.0         & 92.5         \\
BERT+UniKD                            & 84.1          & 90.4          & 89.6         & 93.3         \\ \hline
RoBERTa (single)                     & 85.8          & 91.3          & 90.2         & 93.7         \\
RoBERTa (ensemble)                   & 87.9          & 92.7          & 91.8         & 95.1         \\
RoBERTa+KD-Labeled                           & 86.2          & 91.4          & 90.4         & 93.9         \\
RoBERTa+AE-KD                        & 86.5          & 91.6          & 90.5         & 94.0         \\
RoBERTa+KD-Unlabeled                          & 86.8          & 92.0          & 90.9         & 94.3         \\
RoBERTa+UniKD                         & 87.6          & 92.5          & 91.4         & 94.9         \\ \hline
UniLM (single)                       & 86.8          & 91.7          & 90.1         & 93.8         \\
UniLM (ensemble)                     & 88.4          & 93.0          & 91.8         & 95.3         \\
UniLM+KD-Labeled                             & 87.0          & 91.9          & 90.4         & 94.2         \\
UniLM+AE-KD                          & 87.1          & 92.2          & 90.5         & 94.4         \\
UniLM+KD-Unlabeled                            & 87.5          & 92.4          & 90.8         & 94.6         \\
UniLM+UniKD                           & 88.2          & 92.9          & 91.5         & 95.0         \\ \Xhline{1.5pt}  
\end{tabular}
\end{table}

\subsection{Performance Comparison}

We first verify the effectiveness of UniKD by comparing it with several baseline methods.
We choose the base version of BERT~\cite{devlin2019bert}, RoBERTa~\cite{liu2019roberta}, and UniLM~\cite{bao2020unilmv2} as the basic model.
The methods to be compared include:
(1) single, using a single model for inference;
(2) ensemble, averaging the soft labels predicted by multiple models;
(3) KD-Labeled~\cite{fukuda2017efficient,freitag2017ensemble},  ensemble distillation distillation on labeled data from the averaged soft labels;
(4) AE-KD~\cite{du2020agree}, an adaptive label voting method for ensemble distillation on labeled data;
(5) KD-Unlabeled~\cite{li2019fedmd,sui2020feded}, ensemble distillation distillation on unlabeled data based on averaged predictions on unlabeled data;
(6) UniKD, our proposed unified ensemble knowledge distillation method.
The results on the four datasets are shown in Table~\ref{table.kd}.
We find using ensemble of models can usually greatly improve the accuracy over single models.
This is intuitive because different independent models may encode different knowledge that is complementary to prediction.
However, it also leads to a high computational cost.
The ensemble knowledge distillation methods usually have a better performance than the original single model.
However, the students in all baselines still have a notable gap with the teacher ensemble.
Among them, we find that students distilled on unlabeled data perform better than those distilled on labeled data.
This may be because the student can learn from task labels on labeled data, and the complementary knowledge provided by the teacher on labeled data is insufficient.
Moreover, our UniKD method consistently outperforms other distillation methods with a significant margin ($p<0.05$ in t-test), and can achieve comparable performance with ensemble models.
It shows that UniKD can effectively improve the performance of a single model.
 
 \begin{figure}[!t]
	\centering 
	\includegraphics[width=0.99\linewidth]{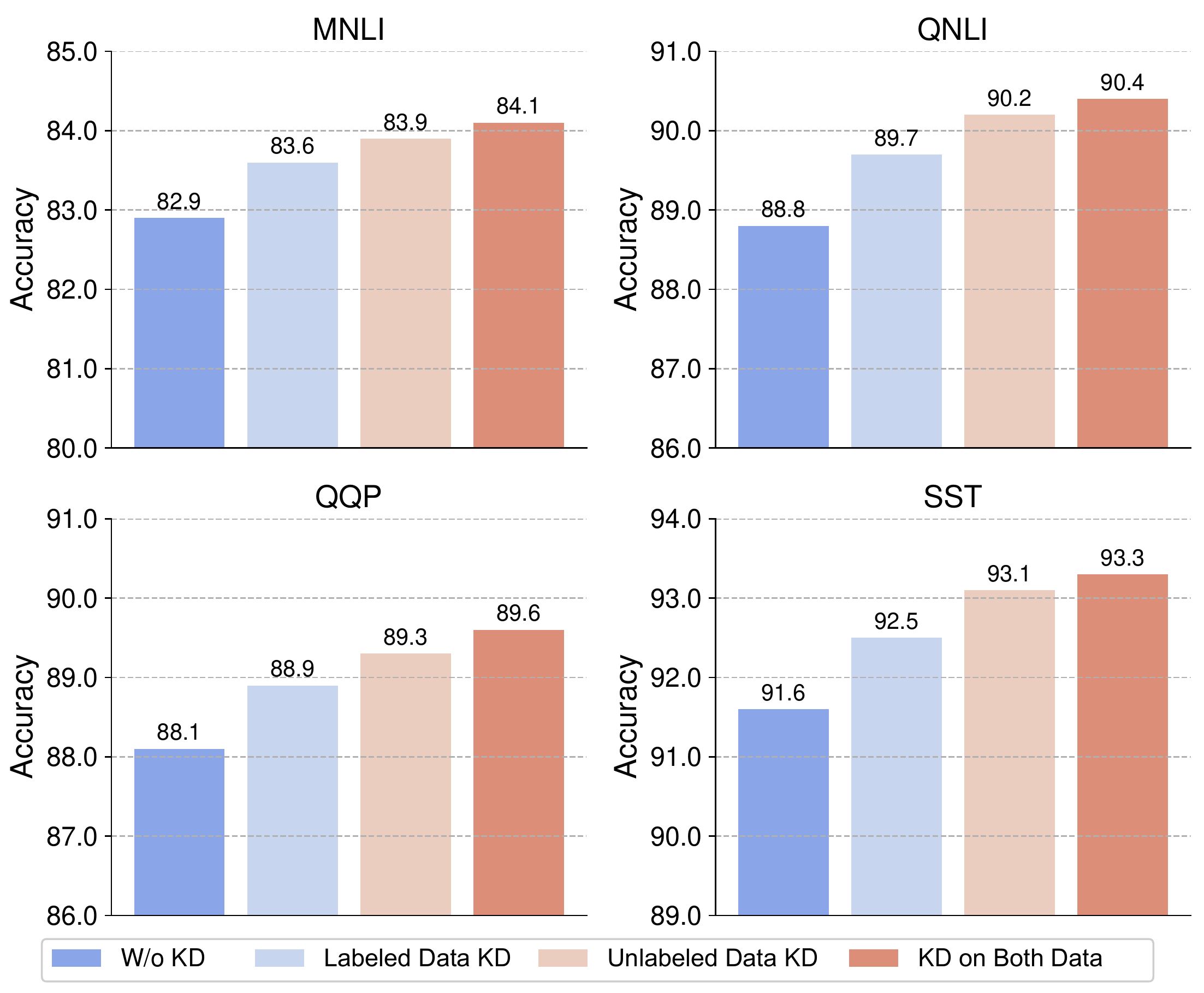} 
\caption{Effectiveness of knowledge distillation on labeled and unlabeled data.}\label{fig.d1}
\end{figure}

\subsection{Ablation Study}
\begin{figure}[!t]
	\centering 
	\includegraphics[width=0.99\linewidth]{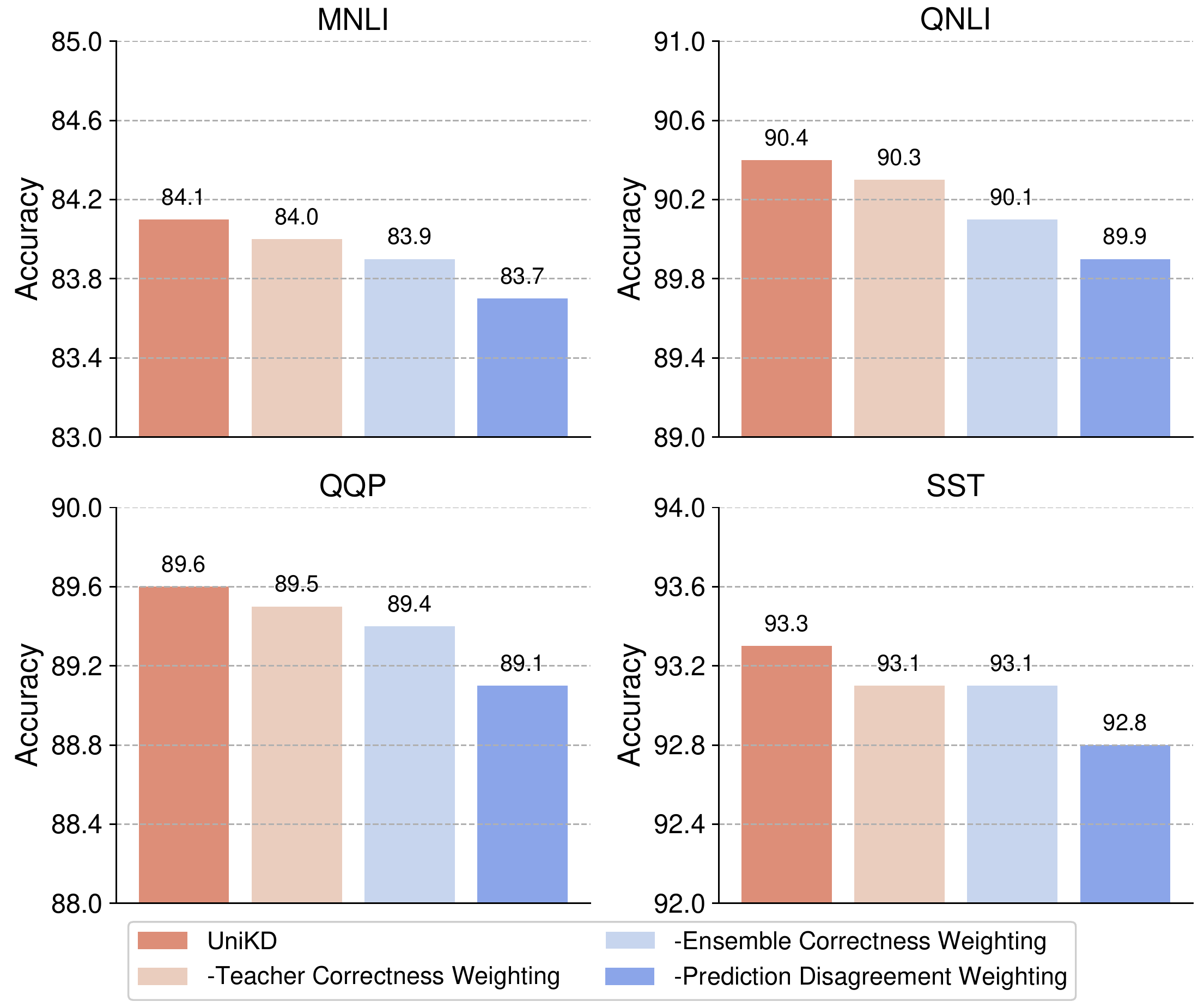} 
\caption{Influence of different weighting mechanisms.}\label{fig.d2}
\end{figure}

Next, we verify the effectiveness of knowledge distillation on labeled and unlabeled data in our method.
We use BERT as the basic model in the following experiments (the experimental results on other basic models show similar patterns, and are omitted due to space limit).
The results are shown in Fig.~\ref{fig.d1}.
We find in our approach knowledge distillation on unlabeled data is also more important than knowledge distillation on labeled data.
It shows the importance of exploiting unlabeled data in knowledge transfer.
Moreover, combining labeled and unlabeled data for knowledge distillation can further improve the student's performance.
This is because distillation on labeled data can reduce the risk of overfitting task labels, and distillation on unlabeled data can help better transfer the knowledge of teacher ensemble.
We then study the influence of different weighting mechanisms in our approach.
The results are shown in Fig.~\ref{fig.d2}.
We find the prediction disagreement weighting mechanism on unlabeled data has the largest contribution. 
This is because it can distinguish the importance of different unlabeled samples, which can help transfer knowledge more effectively.
In addition, both types of correctness weighting methods have some contributions to the performance improvements.
This is because evaluating the teacher models' prediction quality based on task labels can help distill higher-quality knowledge.

\subsection{Hyperparameter Analysis}

\begin{figure}[!t]
	\centering 
	\includegraphics[width=0.99\linewidth]{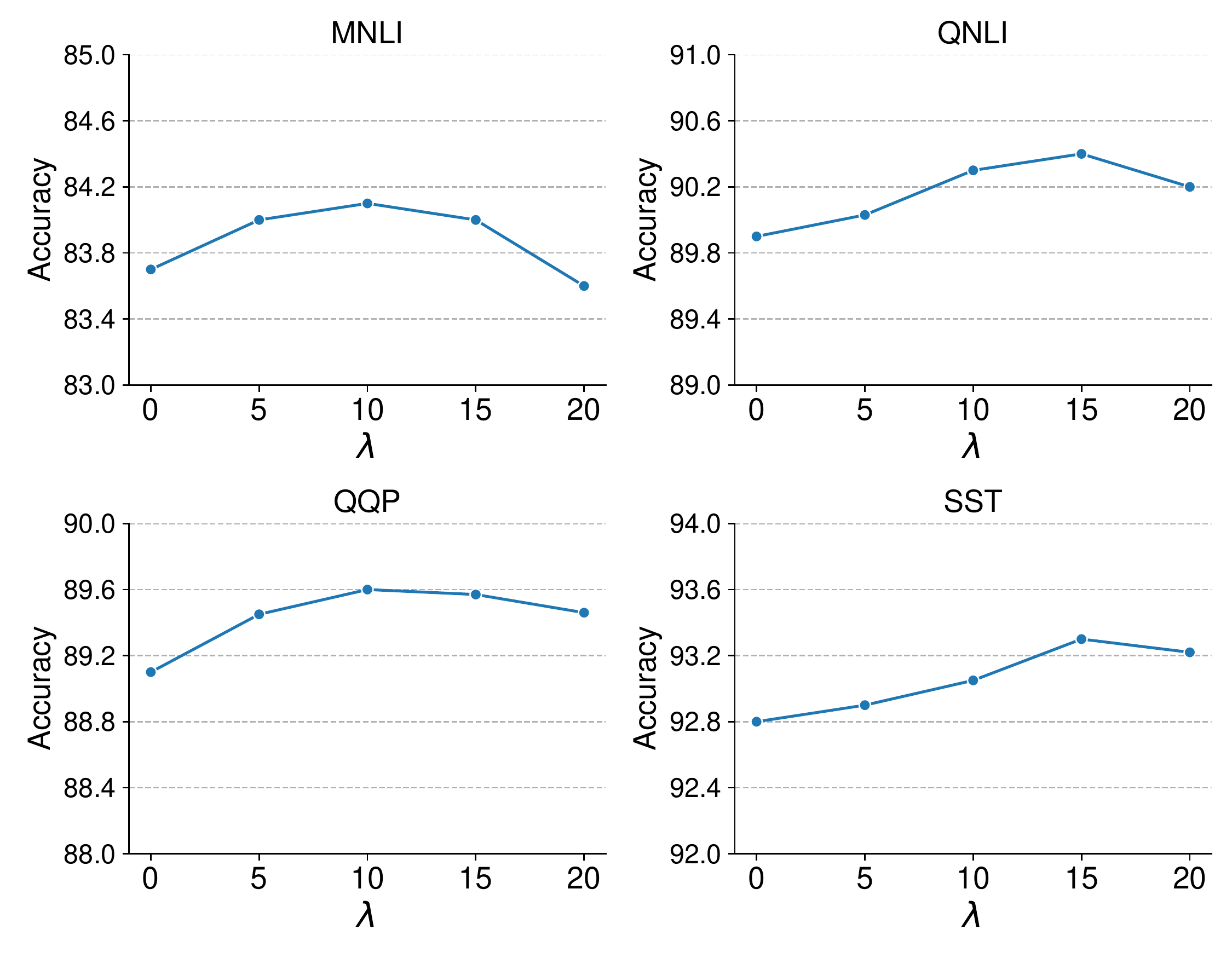} 
\caption{Influence of the weighting  coefficient $\lambda$.}\label{fig.d3}
\end{figure}

Finally, we study the impact of the disagreement weighting coefficient $\lambda$ on the model performance.
The results are shown in Fig.~\ref{fig.d3}.
We find when the value of $\lambda$ is very small, the performance is suboptimal.
This is because the importance of different samples cannot be effectively distinguished.
However, the performance starts to decline when $\lambda$ is too large.
This is because the distillation intensity on unlabeled data becomes too strong, and the model may not fully exploit the supervision signals on labeled data.
Thus, a moderate value of $\lambda$ (e.g., 10 or 15) is more suitable for our approach.

\section{Conclusion}\label{sec:Conclusion}

In this paper, we propose a unified ensemble knowledge distillation method named UniKD, which can effectively transfer useful knowledge from multiple teacher models to a single student model via distilling on both labeled and unlabeled data.
On labeled data, we propose to weight different teachers' soft labels on each sample based on their correctness, and further weight the knowledge distillation loss based on the average correctness of teachers.
On unlabeled data, we propose to use the disagreement of teachers to weight the distillation loss on different samples.
Extensive experiments on four datasets show the effectiveness of our method in boosting the performance of single model,  even approaching the performance of teacher ensemble.

\bibliographystyle{ACM-Reference-Format}
\bibliography{main}


\begin{thebibliography}{37}


\ifx \showCODEN    \undefined \def \showCODEN     #1{\unskip}     \fi
\ifx \showDOI      \undefined \def \showDOI       #1{#1}\fi
\ifx \showISBNx    \undefined \def \showISBNx     #1{\unskip}     \fi
\ifx \showISBNxiii \undefined \def \showISBNxiii  #1{\unskip}     \fi
\ifx \showISSN     \undefined \def \showISSN      #1{\unskip}     \fi
\ifx \showLCCN     \undefined \def \showLCCN      #1{\unskip}     \fi
\ifx \shownote     \undefined \def \shownote      #1{#1}          \fi
\ifx \showarticletitle \undefined \def \showarticletitle #1{#1}   \fi
\ifx \showURL      \undefined \def \showURL       {\relax}        \fi
\providecommand\bibfield[2]{#2}
\providecommand\bibinfo[2]{#2}
\providecommand\natexlab[1]{#1}
\providecommand\showeprint[2][]{arXiv:#2}

\bibitem[\protect\citeauthoryear{Allen-Zhu and Li}{Allen-Zhu and Li}{2020}]%
        {allen2020towards}
\bibfield{author}{\bibinfo{person}{Zeyuan Allen-Zhu} {and}
  \bibinfo{person}{Yuanzhi Li}.} \bibinfo{year}{2020}\natexlab{}.
\newblock \showarticletitle{Towards understanding ensemble, knowledge
  distillation and self-distillation in deep learning}.
\newblock \bibinfo{journal}{\emph{arXiv preprint arXiv:2012.09816}}
  (\bibinfo{year}{2020}).
\newblock


\bibitem[\protect\citeauthoryear{Bao, Dong, Wei, Wang, Yang, Liu, Wang, Gao,
  Piao, Zhou, et~al\mbox{.}}{Bao et~al\mbox{.}}{2020}]%
        {bao2020unilmv2}
\bibfield{author}{\bibinfo{person}{Hangbo Bao}, \bibinfo{person}{Li Dong},
  \bibinfo{person}{Furu Wei}, \bibinfo{person}{Wenhui Wang},
  \bibinfo{person}{Nan Yang}, \bibinfo{person}{Xiaodong Liu},
  \bibinfo{person}{Yu Wang}, \bibinfo{person}{Jianfeng Gao},
  \bibinfo{person}{Songhao Piao}, \bibinfo{person}{Ming Zhou}, {et~al\mbox{.}}}
  \bibinfo{year}{2020}\natexlab{}.
\newblock \showarticletitle{Unilmv2: Pseudo-masked language models for unified
  language model pre-training}. In \bibinfo{booktitle}{\emph{ICML}}. PMLR,
  \bibinfo{pages}{642--652}.
\newblock


\bibitem[\protect\citeauthoryear{Chebotar and Waters}{Chebotar and
  Waters}{2016}]%
        {chebotar2016distilling}
\bibfield{author}{\bibinfo{person}{Yevgen Chebotar} {and}
  \bibinfo{person}{Austin Waters}.} \bibinfo{year}{2016}\natexlab{}.
\newblock \showarticletitle{Distilling Knowledge from Ensembles of Neural
  Networks for Speech Recognition.}. In
  \bibinfo{booktitle}{\emph{Interspeech}}. \bibinfo{pages}{3439--3443}.
\newblock


\bibitem[\protect\citeauthoryear{Cho and Hariharan}{Cho and Hariharan}{2019}]%
        {cho2019efficacy}
\bibfield{author}{\bibinfo{person}{Jang~Hyun Cho} {and}
  \bibinfo{person}{Bharath Hariharan}.} \bibinfo{year}{2019}\natexlab{}.
\newblock \showarticletitle{On the efficacy of knowledge distillation}. In
  \bibinfo{booktitle}{\emph{CVPR}}. \bibinfo{pages}{4794--4802}.
\newblock


\bibitem[\protect\citeauthoryear{Dekel, Gentile, and Sridharan}{Dekel
  et~al\mbox{.}}{2012}]%
        {dekel2012selective}
\bibfield{author}{\bibinfo{person}{Ofer Dekel}, \bibinfo{person}{Claudio
  Gentile}, {and} \bibinfo{person}{Karthik Sridharan}.}
  \bibinfo{year}{2012}\natexlab{}.
\newblock \showarticletitle{Selective sampling and active learning from single
  and multiple teachers}.
\newblock \bibinfo{journal}{\emph{JMLR}} \bibinfo{volume}{13},
  \bibinfo{number}{1} (\bibinfo{year}{2012}), \bibinfo{pages}{2655--2697}.
\newblock


\bibitem[\protect\citeauthoryear{Devlin, Chang, Lee, and Toutanova}{Devlin
  et~al\mbox{.}}{2019}]%
        {devlin2019bert}
\bibfield{author}{\bibinfo{person}{Jacob Devlin}, \bibinfo{person}{Ming-Wei
  Chang}, \bibinfo{person}{Kenton Lee}, {and} \bibinfo{person}{Kristina
  Toutanova}.} \bibinfo{year}{2019}\natexlab{}.
\newblock \showarticletitle{BERT: Pre-training of Deep Bidirectional
  Transformers for Language Understanding}. In
  \bibinfo{booktitle}{\emph{NAACL-HLT}}. \bibinfo{pages}{4171--4186}.
\newblock


\bibitem[\protect\citeauthoryear{Dietterich}{Dietterich}{2000}]%
        {dietterich2000ensemble}
\bibfield{author}{\bibinfo{person}{Thomas~G Dietterich}.}
  \bibinfo{year}{2000}\natexlab{}.
\newblock \showarticletitle{Ensemble methods in machine learning}. In
  \bibinfo{booktitle}{\emph{International workshop on multiple classifier
  systems}}. Springer, \bibinfo{pages}{1--15}.
\newblock


\bibitem[\protect\citeauthoryear{Du, You, Li, Wu, Wang, Qian, and Zhang}{Du
  et~al\mbox{.}}{2020}]%
        {du2020agree}
\bibfield{author}{\bibinfo{person}{Shangchen Du}, \bibinfo{person}{Shan You},
  \bibinfo{person}{Xiaojie Li}, \bibinfo{person}{Jianlong Wu},
  \bibinfo{person}{Fei Wang}, \bibinfo{person}{Chen Qian}, {and}
  \bibinfo{person}{Changshui Zhang}.} \bibinfo{year}{2020}\natexlab{}.
\newblock \showarticletitle{Agree to disagree: Adaptive ensemble knowledge
  distillation in gradient space}.
\newblock \bibinfo{journal}{\emph{NeurIPS}}  \bibinfo{volume}{33}
  (\bibinfo{year}{2020}), \bibinfo{pages}{12345--12355}.
\newblock


\bibitem[\protect\citeauthoryear{Freitag, Al-Onaizan, and Sankaran}{Freitag
  et~al\mbox{.}}{2017}]%
        {freitag2017ensemble}
\bibfield{author}{\bibinfo{person}{Markus Freitag}, \bibinfo{person}{Yaser
  Al-Onaizan}, {and} \bibinfo{person}{Baskaran Sankaran}.}
  \bibinfo{year}{2017}\natexlab{}.
\newblock \showarticletitle{Ensemble distillation for neural machine
  translation}.
\newblock \bibinfo{journal}{\emph{arXiv preprint arXiv:1702.01802}}
  (\bibinfo{year}{2017}).
\newblock


\bibitem[\protect\citeauthoryear{Fukuda, Suzuki, Kurata, Thomas, Cui, and
  Ramabhadran}{Fukuda et~al\mbox{.}}{2017}]%
        {fukuda2017efficient}
\bibfield{author}{\bibinfo{person}{Takashi Fukuda}, \bibinfo{person}{Masayuki
  Suzuki}, \bibinfo{person}{Gakuto Kurata}, \bibinfo{person}{Samuel Thomas},
  \bibinfo{person}{Jia Cui}, {and} \bibinfo{person}{Bhuvana Ramabhadran}.}
  \bibinfo{year}{2017}\natexlab{}.
\newblock \showarticletitle{Efficient Knowledge Distillation from an Ensemble
  of Teachers.}. In \bibinfo{booktitle}{\emph{Interspeech}}.
  \bibinfo{pages}{3697--3701}.
\newblock


\bibitem[\protect\citeauthoryear{Gong, Sharma, Karanam, Wu, Chen, Doermann, and
  Innanje}{Gong et~al\mbox{.}}{2021}]%
        {gong2021ensemble}
\bibfield{author}{\bibinfo{person}{Xuan Gong}, \bibinfo{person}{Abhishek
  Sharma}, \bibinfo{person}{Srikrishna Karanam}, \bibinfo{person}{Ziyan Wu},
  \bibinfo{person}{Terrence Chen}, \bibinfo{person}{David Doermann}, {and}
  \bibinfo{person}{Arun Innanje}.} \bibinfo{year}{2021}\natexlab{}.
\newblock \showarticletitle{Ensemble Attention Distillation for
  Privacy-Preserving Federated Learning}. In \bibinfo{booktitle}{\emph{ICCV}}.
  \bibinfo{pages}{15076--15086}.
\newblock


\bibitem[\protect\citeauthoryear{Ho}{Ho}{1995}]%
        {ho1995random}
\bibfield{author}{\bibinfo{person}{Tin~Kam Ho}.}
  \bibinfo{year}{1995}\natexlab{}.
\newblock \showarticletitle{Random decision forests}. In
  \bibinfo{booktitle}{\emph{ICDAR}}, Vol.~\bibinfo{volume}{1}. IEEE,
  \bibinfo{pages}{278--282}.
\newblock


\bibitem[\protect\citeauthoryear{Kang and Gwak}{Kang and Gwak}{2020}]%
        {kang2020ensemble}
\bibfield{author}{\bibinfo{person}{Jaeyong Kang} {and}
  \bibinfo{person}{Jeonghwan Gwak}.} \bibinfo{year}{2020}\natexlab{}.
\newblock \showarticletitle{Ensemble learning of lightweight deep learning
  models using knowledge distillation for image classification}.
\newblock \bibinfo{journal}{\emph{Mathematics}} \bibinfo{volume}{8},
  \bibinfo{number}{10} (\bibinfo{year}{2020}), \bibinfo{pages}{1652}.
\newblock


\bibitem[\protect\citeauthoryear{Kao, Xie, Lin, and Cheng}{Kao
  et~al\mbox{.}}{2021}]%
        {kao2021specific}
\bibfield{author}{\bibinfo{person}{Wei-Cheng Kao}, \bibinfo{person}{Hong-Xia
  Xie}, \bibinfo{person}{Chih-Yang Lin}, {and} \bibinfo{person}{Wen-Huang
  Cheng}.} \bibinfo{year}{2021}\natexlab{}.
\newblock \showarticletitle{Specific Expert Learning: Enriching Ensemble
  Diversity via Knowledge Distillation}.
\newblock \bibinfo{journal}{\emph{IEEE Transactions on Cybernetics}}
  (\bibinfo{year}{2021}).
\newblock


\bibitem[\protect\citeauthoryear{Kingma and Ba}{Kingma and Ba}{2015}]%
        {kingma2014adam}
\bibfield{author}{\bibinfo{person}{Diederik~P. Kingma} {and}
  \bibinfo{person}{Jimmy Ba}.} \bibinfo{year}{2015}\natexlab{}.
\newblock \showarticletitle{Adam: A Method for Stochastic Optimization}. In
  \bibinfo{booktitle}{\emph{ICLR}}.
\newblock


\bibitem[\protect\citeauthoryear{Krishnan and Tickoo}{Krishnan and
  Tickoo}{2020}]%
        {krishnan2020improving}
\bibfield{author}{\bibinfo{person}{Ranganath Krishnan} {and}
  \bibinfo{person}{Omesh Tickoo}.} \bibinfo{year}{2020}\natexlab{}.
\newblock \showarticletitle{Improving model calibration with accuracy versus
  uncertainty optimization}.
\newblock \bibinfo{journal}{\emph{NeurUPS}}  \bibinfo{volume}{33}
  (\bibinfo{year}{2020}), \bibinfo{pages}{18237--18248}.
\newblock


\bibitem[\protect\citeauthoryear{Li and Wang}{Li and Wang}{2019}]%
        {li2019fedmd}
\bibfield{author}{\bibinfo{person}{Daliang Li} {and} \bibinfo{person}{Junpu
  Wang}.} \bibinfo{year}{2019}\natexlab{}.
\newblock \showarticletitle{Fedmd: Heterogenous federated learning via model
  distillation}.
\newblock \bibinfo{journal}{\emph{arXiv preprint arXiv:1910.03581}}
  (\bibinfo{year}{2019}).
\newblock


\bibitem[\protect\citeauthoryear{Lin, Kong, Stich, and Jaggi}{Lin
  et~al\mbox{.}}{2020}]%
        {lin2020ensemble}
\bibfield{author}{\bibinfo{person}{Tao Lin}, \bibinfo{person}{Lingjing Kong},
  \bibinfo{person}{Sebastian~U Stich}, {and} \bibinfo{person}{Martin Jaggi}.}
  \bibinfo{year}{2020}\natexlab{}.
\newblock \showarticletitle{Ensemble distillation for robust model fusion in
  federated learning}.
\newblock \bibinfo{journal}{\emph{Advances in Neural Information Processing
  Systems}}  \bibinfo{volume}{33} (\bibinfo{year}{2020}),
  \bibinfo{pages}{2351--2363}.
\newblock


\bibitem[\protect\citeauthoryear{Liu, He, Chen, and Gao}{Liu
  et~al\mbox{.}}{2019a}]%
        {liu2019improving}
\bibfield{author}{\bibinfo{person}{Xiaodong Liu}, \bibinfo{person}{Pengcheng
  He}, \bibinfo{person}{Weizhu Chen}, {and} \bibinfo{person}{Jianfeng Gao}.}
  \bibinfo{year}{2019}\natexlab{a}.
\newblock \showarticletitle{Improving multi-task deep neural networks via
  knowledge distillation for natural language understanding}.
\newblock \bibinfo{journal}{\emph{arXiv preprint arXiv:1904.09482}}
  (\bibinfo{year}{2019}).
\newblock


\bibitem[\protect\citeauthoryear{Liu, Ott, Goyal, Du, Joshi, Chen, Levy, Lewis,
  Zettlemoyer, and Stoyanov}{Liu et~al\mbox{.}}{2019b}]%
        {liu2019roberta}
\bibfield{author}{\bibinfo{person}{Yinhan Liu}, \bibinfo{person}{Myle Ott},
  \bibinfo{person}{Naman Goyal}, \bibinfo{person}{Jingfei Du},
  \bibinfo{person}{Mandar Joshi}, \bibinfo{person}{Danqi Chen},
  \bibinfo{person}{Omer Levy}, \bibinfo{person}{Mike Lewis},
  \bibinfo{person}{Luke Zettlemoyer}, {and} \bibinfo{person}{Veselin
  Stoyanov}.} \bibinfo{year}{2019}\natexlab{b}.
\newblock \showarticletitle{Roberta: A robustly optimized bert pretraining
  approach}.
\newblock \bibinfo{journal}{\emph{arXiv preprint arXiv:1907.11692}}
  (\bibinfo{year}{2019}).
\newblock


\bibitem[\protect\citeauthoryear{Mendes-Moreira, Soares, Jorge, and
  Sousa}{Mendes-Moreira et~al\mbox{.}}{2012}]%
        {mendes2012ensemble}
\bibfield{author}{\bibinfo{person}{Joao Mendes-Moreira},
  \bibinfo{person}{Carlos Soares}, \bibinfo{person}{Al{\'\i}pio~M{\'a}rio
  Jorge}, {and} \bibinfo{person}{Jorge Freire~De Sousa}.}
  \bibinfo{year}{2012}\natexlab{}.
\newblock \showarticletitle{Ensemble approaches for regression: A survey}.
\newblock \bibinfo{journal}{\emph{Acm computing surveys (csur)}}
  \bibinfo{volume}{45}, \bibinfo{number}{1} (\bibinfo{year}{2012}),
  \bibinfo{pages}{1--40}.
\newblock


\bibitem[\protect\citeauthoryear{Park and Kwak}{Park and Kwak}{2019}]%
        {park2019feed}
\bibfield{author}{\bibinfo{person}{SeongUk Park} {and} \bibinfo{person}{Nojun
  Kwak}.} \bibinfo{year}{2019}\natexlab{}.
\newblock \showarticletitle{Feed: Feature-level ensemble for knowledge
  distillation}.
\newblock \bibinfo{journal}{\emph{arXiv preprint arXiv:1909.10754}}
  (\bibinfo{year}{2019}).
\newblock


\bibitem[\protect\citeauthoryear{Radosavovic, Doll{\'a}r, Girshick, Gkioxari,
  and He}{Radosavovic et~al\mbox{.}}{2018}]%
        {radosavovic2018data}
\bibfield{author}{\bibinfo{person}{Ilija Radosavovic}, \bibinfo{person}{Piotr
  Doll{\'a}r}, \bibinfo{person}{Ross Girshick}, \bibinfo{person}{Georgia
  Gkioxari}, {and} \bibinfo{person}{Kaiming He}.}
  \bibinfo{year}{2018}\natexlab{}.
\newblock \showarticletitle{Data distillation: Towards omni-supervised
  learning}. In \bibinfo{booktitle}{\emph{CVPR}}. \bibinfo{pages}{4119--4128}.
\newblock


\bibitem[\protect\citeauthoryear{Rajpurkar, Zhang, Lopyrev, and
  Liang}{Rajpurkar et~al\mbox{.}}{2016}]%
        {rajpurkar2016squad}
\bibfield{author}{\bibinfo{person}{Pranav Rajpurkar}, \bibinfo{person}{Jian
  Zhang}, \bibinfo{person}{Konstantin Lopyrev}, {and} \bibinfo{person}{Percy
  Liang}.} \bibinfo{year}{2016}\natexlab{}.
\newblock \showarticletitle{SQuAD: 100,000+ Questions for Machine Comprehension
  of Text}. In \bibinfo{booktitle}{\emph{EMNLP}}. \bibinfo{pages}{2383--2392}.
\newblock


\bibitem[\protect\citeauthoryear{Schapire}{Schapire}{1999}]%
        {schapire1999brief}
\bibfield{author}{\bibinfo{person}{Robert~E Schapire}.}
  \bibinfo{year}{1999}\natexlab{}.
\newblock \showarticletitle{A brief introduction to boosting}. In
  \bibinfo{booktitle}{\emph{IJCAI}}, Vol.~\bibinfo{volume}{99}.
  \bibinfo{pages}{1401--1406}.
\newblock


\bibitem[\protect\citeauthoryear{Socher, Perelygin, Wu, Chuang, Manning, Ng,
  and Potts}{Socher et~al\mbox{.}}{2013}]%
        {socher2013recursive}
\bibfield{author}{\bibinfo{person}{Richard Socher}, \bibinfo{person}{Alex
  Perelygin}, \bibinfo{person}{Jean Wu}, \bibinfo{person}{Jason Chuang},
  \bibinfo{person}{Christopher~D Manning}, \bibinfo{person}{Andrew~Y Ng}, {and}
  \bibinfo{person}{Christopher Potts}.} \bibinfo{year}{2013}\natexlab{}.
\newblock \showarticletitle{Recursive deep models for semantic compositionality
  over a sentiment treebank}. In \bibinfo{booktitle}{\emph{EMNLP}}.
  \bibinfo{pages}{1631--1642}.
\newblock


\bibitem[\protect\citeauthoryear{Sui, Chen, Zhao, Jia, Xie, and Sun}{Sui
  et~al\mbox{.}}{2020}]%
        {sui2020feded}
\bibfield{author}{\bibinfo{person}{Dianbo Sui}, \bibinfo{person}{Yubo Chen},
  \bibinfo{person}{Jun Zhao}, \bibinfo{person}{Yantao Jia},
  \bibinfo{person}{Yuantao Xie}, {and} \bibinfo{person}{Weijian Sun}.}
  \bibinfo{year}{2020}\natexlab{}.
\newblock \showarticletitle{Feded: Federated learning via ensemble distillation
  for medical relation extraction}. In \bibinfo{booktitle}{\emph{EMNLP}}.
  \bibinfo{pages}{2118--2128}.
\newblock


\bibitem[\protect\citeauthoryear{Sun, Tan, Gan, Liu, Zhao, Qin, and Liu}{Sun
  et~al\mbox{.}}{2019}]%
        {sun2019token}
\bibfield{author}{\bibinfo{person}{Hao Sun}, \bibinfo{person}{Xu Tan},
  \bibinfo{person}{Jun-Wei Gan}, \bibinfo{person}{Hongzhi Liu},
  \bibinfo{person}{Sheng Zhao}, \bibinfo{person}{Tao Qin}, {and}
  \bibinfo{person}{Tie-Yan Liu}.} \bibinfo{year}{2019}\natexlab{}.
\newblock \showarticletitle{Token-Level Ensemble Distillation for
  Grapheme-to-Phoneme Conversion}. In \bibinfo{booktitle}{\emph{Interspeech}}.
  \bibinfo{pages}{2115--2119}.
\newblock


\bibitem[\protect\citeauthoryear{Tsoumakas and Vlahavas}{Tsoumakas and
  Vlahavas}{2007}]%
        {tsoumakas2007random}
\bibfield{author}{\bibinfo{person}{Grigorios Tsoumakas} {and}
  \bibinfo{person}{Ioannis Vlahavas}.} \bibinfo{year}{2007}\natexlab{}.
\newblock \showarticletitle{Random k-labelsets: An ensemble method for
  multilabel classification}. In \bibinfo{booktitle}{\emph{European conference
  on machine learning}}. Springer, \bibinfo{pages}{406--417}.
\newblock


\bibitem[\protect\citeauthoryear{Walawalkar, Shen, and Savvides}{Walawalkar
  et~al\mbox{.}}{2020}]%
        {walawalkar2020online}
\bibfield{author}{\bibinfo{person}{Devesh Walawalkar},
  \bibinfo{person}{Zhiqiang Shen}, {and} \bibinfo{person}{Marios Savvides}.}
  \bibinfo{year}{2020}\natexlab{}.
\newblock \showarticletitle{Online ensemble model compression using knowledge
  distillation}. In \bibinfo{booktitle}{\emph{ECCV}}. Springer,
  \bibinfo{pages}{18--35}.
\newblock


\bibitem[\protect\citeauthoryear{Wang, Singh, Michael, Hill, Levy, and
  Bowman}{Wang et~al\mbox{.}}{2018}]%
        {wang2018glue}
\bibfield{author}{\bibinfo{person}{Alex Wang}, \bibinfo{person}{Amanpreet
  Singh}, \bibinfo{person}{Julian Michael}, \bibinfo{person}{Felix Hill},
  \bibinfo{person}{Omer Levy}, {and} \bibinfo{person}{Samuel Bowman}.}
  \bibinfo{year}{2018}\natexlab{}.
\newblock \showarticletitle{GLUE: A Multi-Task Benchmark and Analysis Platform
  for Natural Language Understanding}. In
  \bibinfo{booktitle}{\emph{BlackboxNLP}}. \bibinfo{pages}{353--355}.
\newblock


\bibitem[\protect\citeauthoryear{Wang, Li, Liu, Wu, and Zhu}{Wang
  et~al\mbox{.}}{2020}]%
        {wang2020distilling}
\bibfield{author}{\bibinfo{person}{Zirui Wang}, \bibinfo{person}{Bo Li},
  \bibinfo{person}{Naihao Liu}, \bibinfo{person}{Bangyu Wu}, {and}
  \bibinfo{person}{Xu Zhu}.} \bibinfo{year}{2020}\natexlab{}.
\newblock \showarticletitle{Distilling knowledge from an ensemble of
  convolutional neural networks for seismic fault detection}.
\newblock \bibinfo{journal}{\emph{IEEE Geoscience and Remote Sensing Letters}}
  (\bibinfo{year}{2020}).
\newblock


\bibitem[\protect\citeauthoryear{Williams, Nangia, and Bowman}{Williams
  et~al\mbox{.}}{2018}]%
        {williams2018broad}
\bibfield{author}{\bibinfo{person}{Adina Williams}, \bibinfo{person}{Nikita
  Nangia}, {and} \bibinfo{person}{Samuel Bowman}.}
  \bibinfo{year}{2018}\natexlab{}.
\newblock \showarticletitle{A Broad-Coverage Challenge Corpus for Sentence
  Understanding through Inference}. In \bibinfo{booktitle}{\emph{NAACL}}.
  \bibinfo{pages}{1112--1122}.
\newblock


\bibitem[\protect\citeauthoryear{Wu, Wu, and Huang}{Wu et~al\mbox{.}}{2021}]%
        {wu2021one}
\bibfield{author}{\bibinfo{person}{Chuhan Wu}, \bibinfo{person}{Fangzhao Wu},
  {and} \bibinfo{person}{Yongfeng Huang}.} \bibinfo{year}{2021}\natexlab{}.
\newblock \showarticletitle{One Teacher is Enough? Pre-trained Language Model
  Distillation from Multiple Teachers}. In \bibinfo{booktitle}{\emph{ACL
  Findings}}. \bibinfo{pages}{4408--4413}.
\newblock


\bibitem[\protect\citeauthoryear{Xu, Qiu, Zhou, and Huang}{Xu
  et~al\mbox{.}}{2020}]%
        {xu2020improving}
\bibfield{author}{\bibinfo{person}{Yige Xu}, \bibinfo{person}{Xipeng Qiu},
  \bibinfo{person}{Ligao Zhou}, {and} \bibinfo{person}{Xuanjing Huang}.}
  \bibinfo{year}{2020}\natexlab{}.
\newblock \showarticletitle{Improving bert fine-tuning via self-ensemble and
  self-distillation}.
\newblock \bibinfo{journal}{\emph{arXiv preprint arXiv:2002.10345}}
  (\bibinfo{year}{2020}).
\newblock


\bibitem[\protect\citeauthoryear{Yuan, Tay, Li, Wang, and Feng}{Yuan
  et~al\mbox{.}}{2020}]%
        {yuan2020revisiting}
\bibfield{author}{\bibinfo{person}{Li Yuan}, \bibinfo{person}{Francis~EH Tay},
  \bibinfo{person}{Guilin Li}, \bibinfo{person}{Tao Wang}, {and}
  \bibinfo{person}{Jiashi Feng}.} \bibinfo{year}{2020}\natexlab{}.
\newblock \showarticletitle{Revisiting knowledge distillation via label
  smoothing regularization}. In \bibinfo{booktitle}{\emph{CVPR}}.
  \bibinfo{pages}{3903--3911}.
\newblock


\bibitem[\protect\citeauthoryear{Zhu, Gong, et~al\mbox{.}}{Zhu
  et~al\mbox{.}}{2018}]%
        {zhu2018knowledge}
\bibfield{author}{\bibinfo{person}{Xiatian Zhu}, \bibinfo{person}{Shaogang
  Gong}, {et~al\mbox{.}}} \bibinfo{year}{2018}\natexlab{}.
\newblock \showarticletitle{Knowledge distillation by on-the-fly native
  ensemble}.
\newblock \bibinfo{journal}{\emph{NIPS}}  \bibinfo{volume}{31}
  (\bibinfo{year}{2018}).
\newblock


\end{thebibliography}

\end{document}